\documentclass[letterpaper, 10 pt, conference]{ieeeconf}  % Comment this line out if you need a4paper

\IEEEoverridecommandlockouts                              % This command is only needed if 
\usepackage{xcolor}

\usepackage{graphics} % for pdf, bitmapped graphics files
\usepackage{epsfig} % for postscript graphics files
\usepackage{censor}  % censor
\usepackage{amsmath} % assumes amsmath package installed
\usepackage{amssymb}  % assumes amsmath package installed
\usepackage{float}  
\usepackage{subcaption} % for subfigure captions

\usepackage{booktabs}
\usepackage{tabularx}
\usepackage{multirow}
\newcolumntype{Y}{>{\centering\arraybackslash}X} % Create a centered column type that shares remaining width equally

\usepackage{pifont} % for red cross
\usepackage[dvipsnames]{xcolor} % for extra colors
\usepackage{colortbl} % used by tables
\colorlet{colorFst}{Green!25}       % first
\colorlet{colorSnd}{SpringGreen!45} % second
\colorlet{colorTrd}{Yellow!30}      % third
\newcommand{\fst}{\cellcolor{colorFst}\bf}   % first
\newcommand{\nd}{\cellcolor{colorSnd}}      % second
\newcommand{\rd}{\cellcolor{colorTrd}}      % third
\newcommand{\greencheck}{{\color{Green}\checkmark}}
\newcommand{\redx}{{\color{red}\ding{55}}}

\title{\LARGE \bf
Sensor Configuration Matters: A Systematic Evaluation of Multimodal SLAM on Quadruped Robots
}

\author{Roberto Corlito$^{1}$, Fabian Schmidt$^{2,3}$, Nils Seibert$^{1}$,  Markus Enzweiler$^{2}$, Abhinav Valada$^{3}$ and Arne Roennau$^{1}$% <-this % stops a space
\thanks{$^{1}$Machine Intelligence and Robotics Lab, Karlsruhe Institute of Technology (KIT), Germany.
}%
\thanks{$^{2}$ Institute for Intelligent Systems, Esslingen University of Applied Sciences, Germany.}%
\thanks{$^{3}$ Department of Computer Science, University of Freiburg, Germany.}%
}

\begin{document}

\maketitle
\thispagestyle{empty}
\pagestyle{empty}

%%%%%%%%%%%%%%%%%%%%%%%%%%%%%%%%%%%%%%%%%%%%%%%%%%%%%%%%%%%%%%%%%%%%%%%%%%%%%%%%
\begin{abstract}

Autonomous navigation of quadrupedal robots in diverse environments fundamentally relies on resilient Simultaneous Localization and Mapping (SLAM). While visual-inertial SLAM has matured across wheeled, handheld, and aerial platforms, a critical evaluation gap remains regarding how hardware-level sensor configurations affect performance under the aggressive dynamics of legged locomotion. Quadrupeds introduce distinct embodiment-induced sensory challenges, including foot-impact shocks, high-frequency mechanical vibrations, and rapid angular rotations, which degrade standard perception pipelines. 
To address this gap, we present a systematic evaluation of state-of-the-art visual, visual-inertial, and LiDAR-visual-inertial SLAM methods using the GrandTour dataset recorded on an ANYmal D quadruped. We isolate and quantify the impacts of camera modalities, shutter techniques, and inertial sensor tiers, analyzing their trade-offs across localization accuracy, algorithmic robustness, and computational resource utilization. 
Our empirical findings demonstrate that hardware selection has substantial influence on system resilience: stereo configurations consistently outperform monocular and RGB-D modalities, global shutter cameras significantly mitigate motion-induced tracking failures compared to rolling shutter cameras, and, crucially, standard inertial integration can degrade the performance of primarily vision-based frameworks under harsh legged locomotion. These insights additionally offer concrete design guidelines for tailoring custom sensor payloads to achieve dependable perception on agile legged systems.

% To close this gap, we evaluate State-of-the-Art state vision-centric methods on the GrandTour dataset recorded with the ANYmal D quadruped robot. We analyze the effect of different sensor configurations on the performance of those methods on robustness, accuracy and resource usage. 
%\TODO{AR: extremfall ist quadrupeds, wir schauen wie kann man vslam zum laufen bringen bsp grandtour, größerer scope gestalten und dann quadrupeds als extremfall vorstellen}

% We find that the choice of sensor configurations has substantial influence on robustness and accuracy. Our findings can enable the design of custom robots and sensor payloads for more accurate and robust perception capabilities.

% Visual-inertial methods for state estimation have been evaluated on a vast set of platforms. However, there is still a lack of evaluation of different Visual-Inertial configurations on walking robots. Walking robots present unique challenges in vision-centric state estimation because of their dynamic movement.

\end{abstract}

%%%%%%%%%%%%%%%%%%%%%%%%%%%%%%%%%%%%%%%%%%%%%%%%%%%%%%%%%%%%%%%%%%%%%%%%%%%%%%%%
\section{INTRODUCTION}

Modern field robotics increasingly targets unstructured outdoor environments such as disaster zones, industrial facilities, and rugged terrains. 
Traditional unmanned ground vehicles (UGVs), while reliable on mostly flat ground, face substantial traversal constraints on steep inclines, muddy topography, or discrete obstacles like stairs.
To overcome these limitations, legged platforms like the ANYmal D quadruped~\cite{hutter2016anymal} utilize stepping capabilities to negotiate uneven pathways and traverse complex obstacles where conventional robots struggle~\cite{tranzatto2022cerberus}.

Reliable legged autonomy requires resilient, real-time simultaneous localization and mapping (SLAM) to anchor path planning, obstacle avoidance, and state feedback loops~\cite{miki2022learning}.
Without precise localization and environmental mapping, a quadruped lacks the spatial awareness required to safely navigate complex hazards and execute autonomous missions in unstructured environments~\cite{frey2023fast}.

Deploying existing SLAM frameworks onto quadrupedal platforms reveals a distinct algorithmic disconnect. 
Most state-of-the-art algorithms are designed and validated using smooth motion profiles from handheld sensors, drones, or wheeled platforms~\cite{campos2021orbslam3, labbe2019rtabmap}. 
Legged locomotion introduces a hostile sensory environment characterized by intermittent foot contacts, ground slippage, sharp angular velocity spikes, and violent mechanical shocks. 
These locomotion dynamics degrade standard visual tracking by causing severe motion blur and geometric distortions that subsequently disrupt frame-to-frame data association, resulting in poor feature matching that degrades pose estimation accuracy and induces rapid estimator drift~\cite{wisth2022vilens}.

Investigating these hardware vulnerabilities has been constrained by a scarcity of legged robot datasets that allow researchers to systematically vary sensor specifications under realistic physical disturbances~\cite{chaney2023m3ed, nwankwo2024envodat, yao2024tail, jiao2022fusionportable}. 
The GrandTour dataset~\cite{frey2026grandtour} addresses this limitation by offering comprehensive multi-sensor trajectories captured across diverse indoor and outdoor environments using the Boxi payload~\cite{frey2025boxi}. 
While the original GrandTour benchmark evaluates various SLAM baselines under fixed sensor conditions, it does not isolate sensor-specific hardware effects. This paper complements those algorithmic evaluations by leveraging the dataset to systematically alter the underlying hardware configurations, isolating camera modalities, shutter mechanisms, and inertial tiers to identify the optimal sensor baseline for legged autonomy.

This work addresses two main research objectives. 
First, across optimization-based, data-driven, and multi-modal SLAM methods, we isolate and quantify the specific performance impacts of varying camera modalities, shutter technologies, and inertial sensor tiers within visual, visual-inertial, and LiDAR-visual-inertial configurations. 
Second, using the optimal hardware baseline identified for each method, we cross evaluate the frameworks to pinpoint the most robust and best performing configuration while explicitly weighing tracking accuracy against computational resource utilization under quadruped locomotion.

In summary, this work provides three core insights for legged robot design. First, stereo hardware setups consistently yield superior tracking performance compared to monocular and RGB-D configurations. Second, global shutter exposure mechanisms mitigate locomotion induced motion blur to dramatically expand tracking robustness over rolling shutter alternatives. Finally, tightly coupled inertial data integration can counterintuitively degrade tracking resilience for optimization-based frameworks like ORB-SLAM3~\cite{campos2021orbslam3} and RTAB-Map~\cite{labbe2019rtabmap} when subjected to the high frequency mechanical shocks of quadrupedal locomotion.

%%%%%%%%%%%%%%%%%%%%%%%%%%%%%%%%%%%%%%%%%%%%%%%%%%%%%%%%%%%%%%%%%%%%%%%%%%%%%%%%
\section{RELATED WORK}
\label{sec:related work}

Systematic benchmarking is critical for identifying the performance limits of multi-sensor fusion frameworks across complex operational domains. This section contextualizes this work within the SLAM benchmarking literature, moving from environment-focused datasets to platform-driven sensor specification evaluations.

\subsection{Standard and Unstructured Environment Benchmarks}

Visual SLAM progress has historically relied on standardized public datasets. 
Legacy benchmarks like EuRoC MAV~\cite{burri2016euroc}, KITTI~\cite{geiger2012kitti}, and TUM RGB-D~\cite{sturm2012tumrgbd} serve as gold standards for validating tracking accuracy, but they primarily evaluate state estimation across structured geometries or clean motion profiles such as rigid urban roads, indoor laboratories, or stable aerial drone trajectories.

To bypass these limitations, recent literature addresses complex, unstructured environments. 
The TartanAir dataset~\cite{wang2020tartanair} provides simulated outdoor environments with challenging lighting and weather variations, while the TUM-VI benchmark~\cite{schubert2018tumvi} evaluates tracking across long trajectories in mixed indoor and outdoor settings. 
Further, prior benchmarks investigated the performance gradients of traditional visual SLAM and visual-inertial odometry frameworks within variable natural settings like gardens and parks~\cite{schmidt2024vislam, schmidt2025vislam, schmidt2025rover}, isolating the impact of framework architectures and hardware footprints under dense foliage, canopy coverage, and dynamic illumination. 
This evaluation paradigm has since been extended to neural scene representations~\cite{schmidt2024nerf}.

Despite valuable insights into environmental and algorithmic constraints, these benchmarks feature heterogeneous platform dynamics ranging from agile micro-aerial flights and continuous car odometry to smooth handheld tracks. 
Crucially, none of these diverse carrier profiles capture the extreme mechanical vibrations and high-frequency impact shocks inherent to legged locomotion, which introduce aggressive ego-motion and jitter that can instantly invalidate standard tracking assumptions.

\subsection{Legged Platforms and Sensor Specification Benchmarks}

To address platform dynamics, specialized legged datasets have recently emerged. Datasets like SubT-MRS~\cite{zhao2024subt}, EnvoDat~\cite{nwankwo2024envodat}, TAIL~\cite{yao2024tail}, and FusionPortable~\cite{jiao2022fusionportable, wei2024fusionportablev2} supply multi-modal sensor streams from quadrupeds operating across diverse indoor, underground, and outdoor domains. 
Crucially, the GrandTour dataset~\cite{frey2026grandtour} provides large-scale trajectories featuring an ANYmal quadruped under extreme weather and lighting. 
However, their baseline benchmarks primarily focus on confirming localizability using a single, fixed high-end sensor suite, leaving it unclear how tracking resilience fluctuates for varying hardware specifications.

Isolating the impact of sensor configurations across a uniform pipeline remains an open challenge, predominantly studied on wheeled platforms. 
For instance, the ROVER dataset and benchmark~\cite{schmidt2025rover} mapped multi-modal camera configurations on an UGV, but the evaluation was restricted to vision-only methods in semi-structured outdoor settings. 
Further, a wheeled chassis filters out high-frequency micro-shocks and angular velocity spikes due to its continuous wheel-ground contact dynamics and suspension damping. 
Consequently, a configuration which is robust on a smooth UGV may fail under the unmodeled hardware-level data degradation caused by legged locomotion.

To bridge these gaps, this work adapts the hardware-centric benchmarking methodology of~\cite{schmidt2025rover} to the extreme dynamics of legged platforms while expanding environmental and modal diversity. 
Utilizing the GrandTour dataset, we systematically vary sensor specifications across interconnected indoor and outdoor tracks. 
We extend the algorithmic scope beyond vision-only methods by evaluating four state-of-the-art frameworks: RTAB-Map~\cite{labbe2019rtabmap}, ORB-SLAM3~\cite{campos2021orbslam3}, FAST-LIVO2~\cite{zheng2024fastlivo2}, and DPV-SLAM~\cite{lipson2024dpvslam}. 
Specifically, we evaluate camera modalities (monocular, stereo, RGB-D), quantify tracking degradation from shutter artifacts (global vs. rolling shutter) under rapid ego-motion, and benchmark how internal sensor performance tiers (industrial-grade vs. tactical-grade IMUs) impact multi-sensor fusion robustness.

%%%%%%%%%%%%%%%%%%%%%%%%%%%%%%%%%%%%%%%%%%%%%%%%%%%%%%%%%%%%%%%%%%%%%%%%%%%%%%%%
\section{Experimental Setup}
\label{sec:method}

Evaluating SLAM frameworks under diverse operational conditions is critical for characterizing their real-world performance limits, tracking vulnerabilities, and identifying failure modes. 
To establish a rigorous and reproducible baseline under the aggressive dynamics of legged locomotion, this section defines the SLAM method selection criteria, characterizes the hardware of the multi-sensor payload, and details the environmental profiles of the target evaluation tracks.

\subsection{SLAM Method Selection}

The choice of evaluated frameworks is driven by explicit operational criteria, requiring an open-source implementation (for reproducibility), real-time processing capability, and a prioritization of global loop closure features for long-term consistency. 
To ensure broad algorithmic diversity under these prerequisites, we select state-of-the-art frameworks categorized into three distinct methodological classes: traditional optimization-based SLAM, data-driven pipelines, and multi-sensor fusion utilizing active range sensors.% FAST-LIVO2 serves as a reference baseline due to its strong performance in the GrandTour benchmark, while ORB-SLAM3, RTAB-Map, and DPV-SLAM extend the evaluation.
%While FAST-LIVO2 is the only framework shared with the original GrandTour evaluation, where it ranked second among all LIO and LIVO approaches, the inclusion of ORB-SLAM3, RTAB-Map, and DPV-SLAM extends the benchmark to widely used visual and visual-inertial SLAM systems.
%Here, only FAST-LIVO2 was used in the GrandTour evaluation, which was second among LIO and LIVO approaches.

Representing the first class of traditional optimization-based approaches, ORB-SLAM3 is a SLAM system that natively supports monocular, stereo, and RGB-D modalities, with each configuration capable of tightly coupling with inertial measurements.
It employs a sparse, feature-based tracking frontend and relies on local bundle adjustment (BA) for local state estimation. 
For global consistency, it utilizes a bag-of-words (BoW)~\cite{galvez2012bow} approach for loop detection, combined with pose-graph optimization (PGO) and global BA for loop correction.

Also within the optimization-based class, RTAB-Map is a modular SLAM library supporting multiple distinct feature detection and description methods that can optionally integrate LiDAR measurements. 
For this evaluation, the framework is restricted to visual and visual-inertial tracking configurations utilizing the stereo camera payload. 
Similar to other traditional approaches, it executes local BA for frame-to-frame state estimation, while its global backend handles loop correction through BoW place recognition and PGO.

Many learning-based state estimators are excluded from this study due to prohibitive computational costs, a lack of real-time tracking execution, or an absence of global backend correction. 
To evaluate a candidate within the data-driven class, DPV-SLAM is selected as it delivers deterministic real-time tracking and low GPU memory overhead, despite being structurally restricted to monocular configurations. 
DPV-SLAM leverages dense optical flow for camera tracking via the DPVO~\cite{teed2023dpvo} frontend. 
For state estimation, it utilizes local BA, which is coupled with a BoW loop detection scheme, PGO, and global BA for loop correction.

Finally, we evaluate FAST-LIVO2 to investigate the third class of methods, focusing on how visual tracking specifications impact a tightly coupled filter backend when paired with active range sensors. 
FAST-LIVO2 fuses dense LiDAR measurements, inertial data, and camera streams via an error-state iterated Kalman filter without loop closure capabilities.
Dense LiDAR geometry provides a stable tracking anchor, allowing us to isolate whether hardware-level visual artifacts such as rolling-shutter distortions degrade state estimation even in a robust multi-modal pipeline.
% Utilizing dense, full-field-of-view LiDAR geometry provides a stable tracking anchor, allowing us to isolate and observe whether hardware-level visual artifacts, such as rolling-shutter geometric distortions, can degrade state estimation performance even within a highly robust multi-modal pipeline.

\begin{table*}
\caption{Selected missions from the GrandTour dataset and their descriptive environmental attributes. }
\label{tab:mission_selection}
\setlength{\tabcolsep}{5pt}
\begin{center}
\begin{tabular}{llccclll}
\toprule
\textbf{Mission} & \textbf{ID} & \textbf{Duration [s]} & \textbf{Distance [m]} & \textbf{Loops} & \textbf{Environment} & \textbf{Illumination} & \textbf{Terrain and Conditions} \\
\midrule
SNOW-3  & M10 & 236 & 145.3 & \greencheck & Outdoor Mountain      & Sunlight & Snow, Slopes \\
GRI-1   & M13 & 455 & 266.8 & \greencheck & Outdoor Urban         & Sunlight & Grass, Pavement, Stairs, People, Cars \\
PIL-2   & M19 & 409 & 273.6 & \greencheck* & Outdoor Mountain      & Sunlight & Gravel, Dirt Trail, Stairs, People \\
HEAP-1  & M24 & 290 & 162.0 & \redx & Outdoor Industrial    & Cloudy   & Mud, Grass, Standing Water \\
LEE-1   & M34 & 529 & 331.8 & \greencheck & Underground / Outdoor & Cloudy   & Industrial, Pavement, Dim Lighting, People \\
LEICA-1 & M42 & 310 & 152.7 & \greencheck & Indoor / Outdoor      & Sunlight & Urban Pavement, People, Cars \\
SBB-1   & M44 & 502 & 311.3 & \greencheck* & Outdoor Industrial    & Sunlight & Gravel, Soil, Railway Tracks \\
\bottomrule
\end{tabular}
\end{center}
{\footnotesize
      \hspace{1em} Fields marked with * include a reverse pass loop, which structurally is a loop, but visually degraded and hard to detect.}
\end{table*}

\begin{figure*}
    \centering
    \includegraphics[width=0.1375\textwidth]{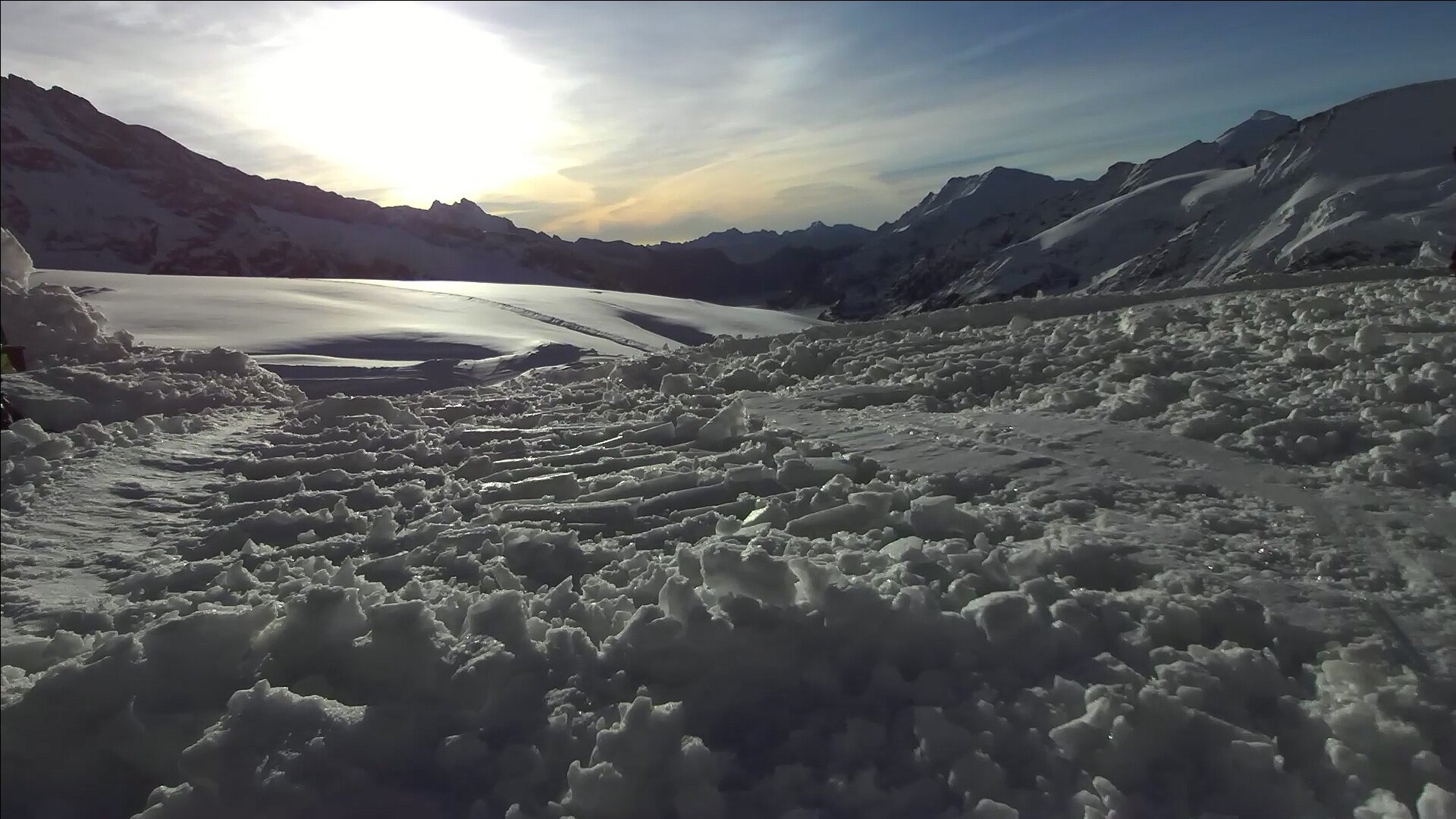}
    \includegraphics[width=0.1375\textwidth]{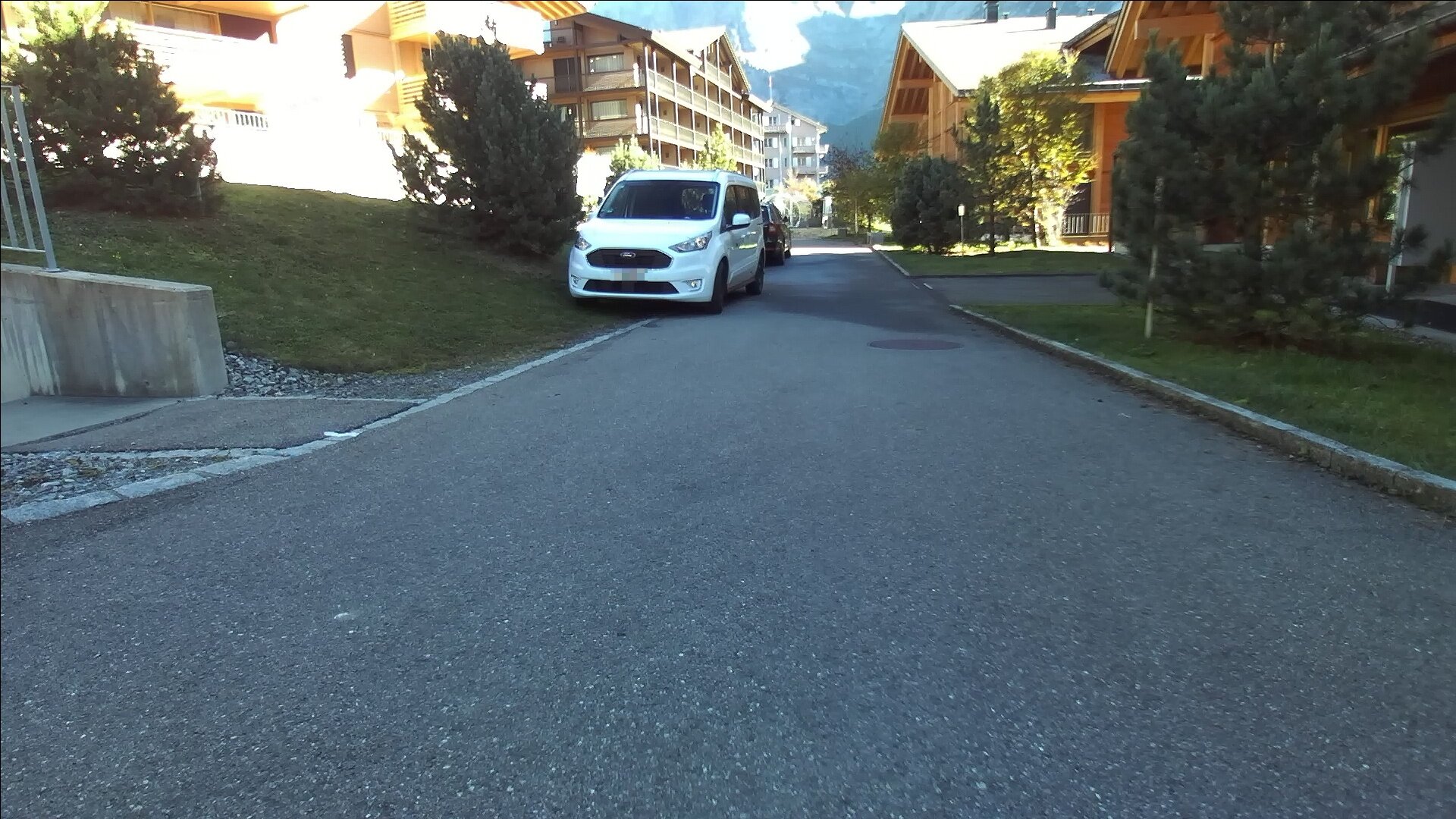}
    \includegraphics[width=0.1375\textwidth]{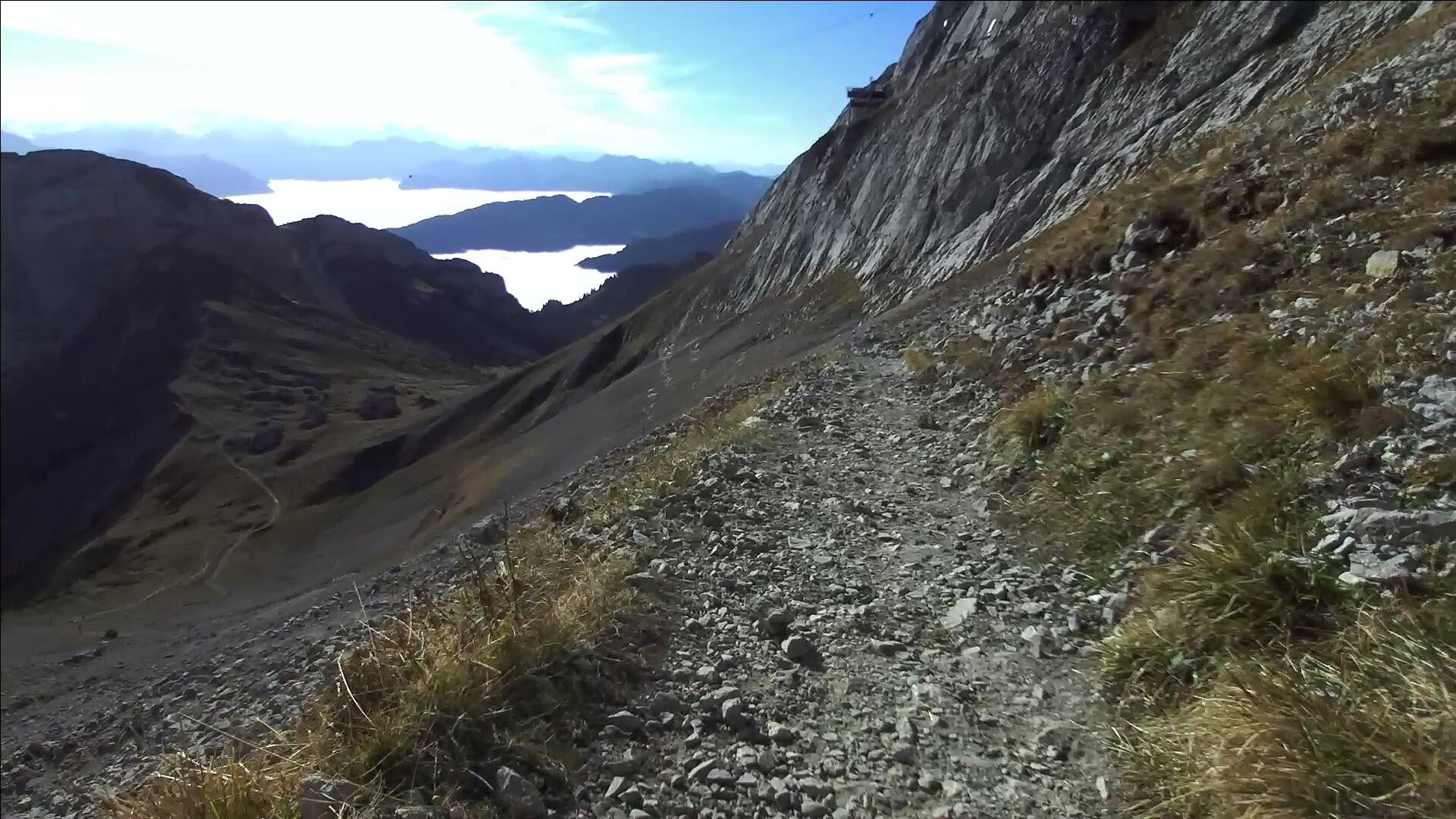}
    \includegraphics[width=0.1375\textwidth]{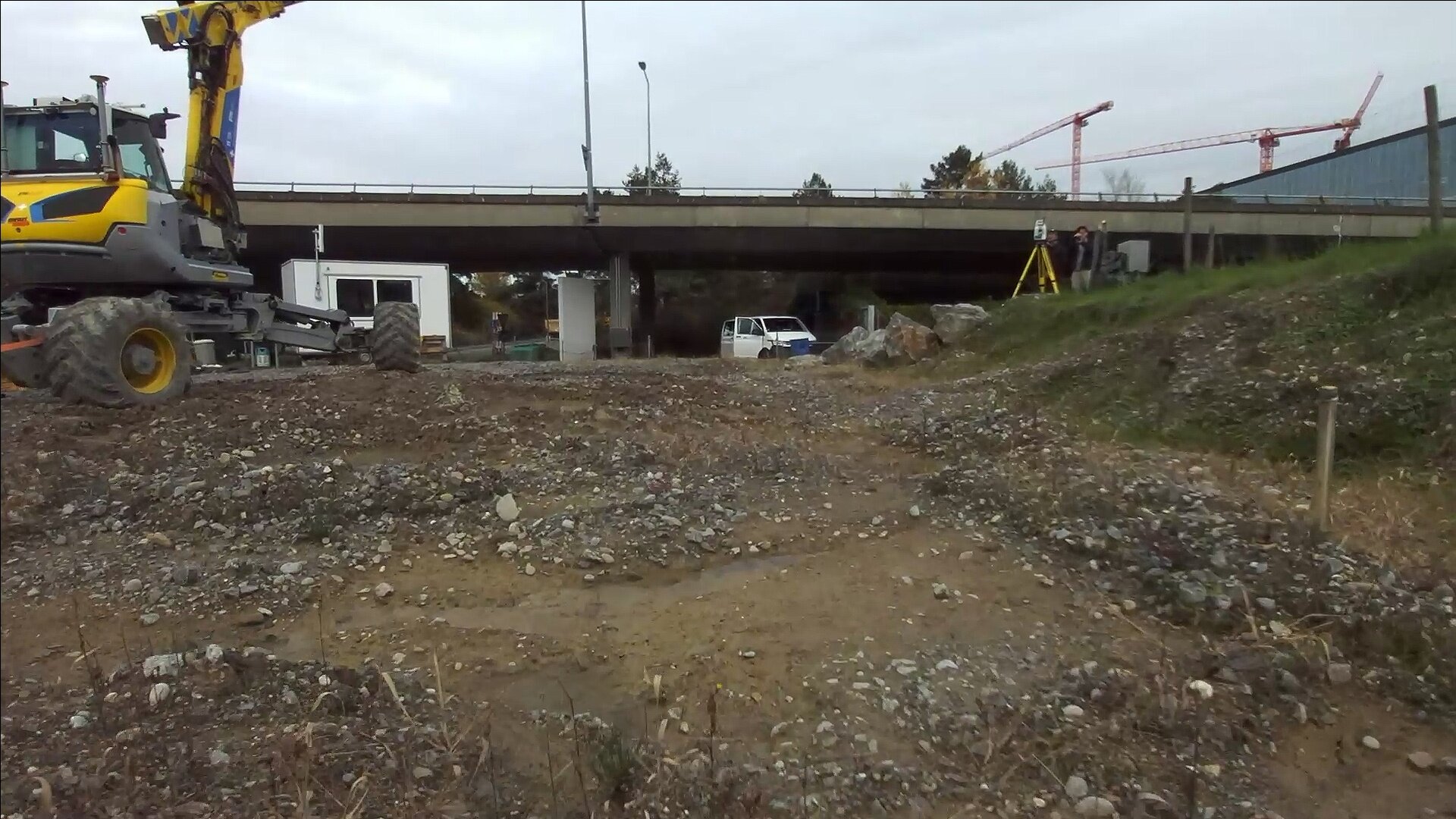}
    \includegraphics[width=0.1375\textwidth]{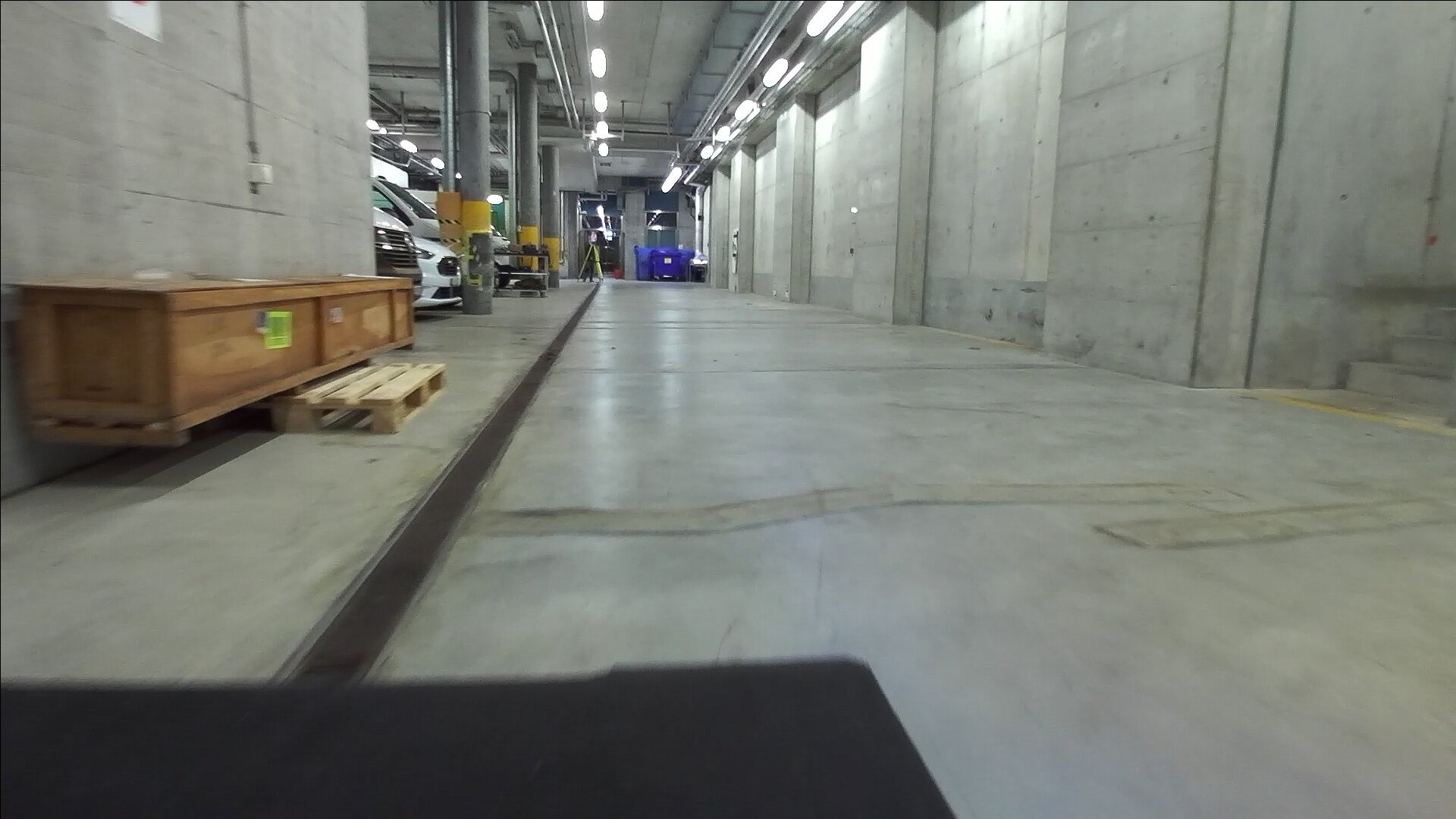}
    \includegraphics[width=0.1375\textwidth]{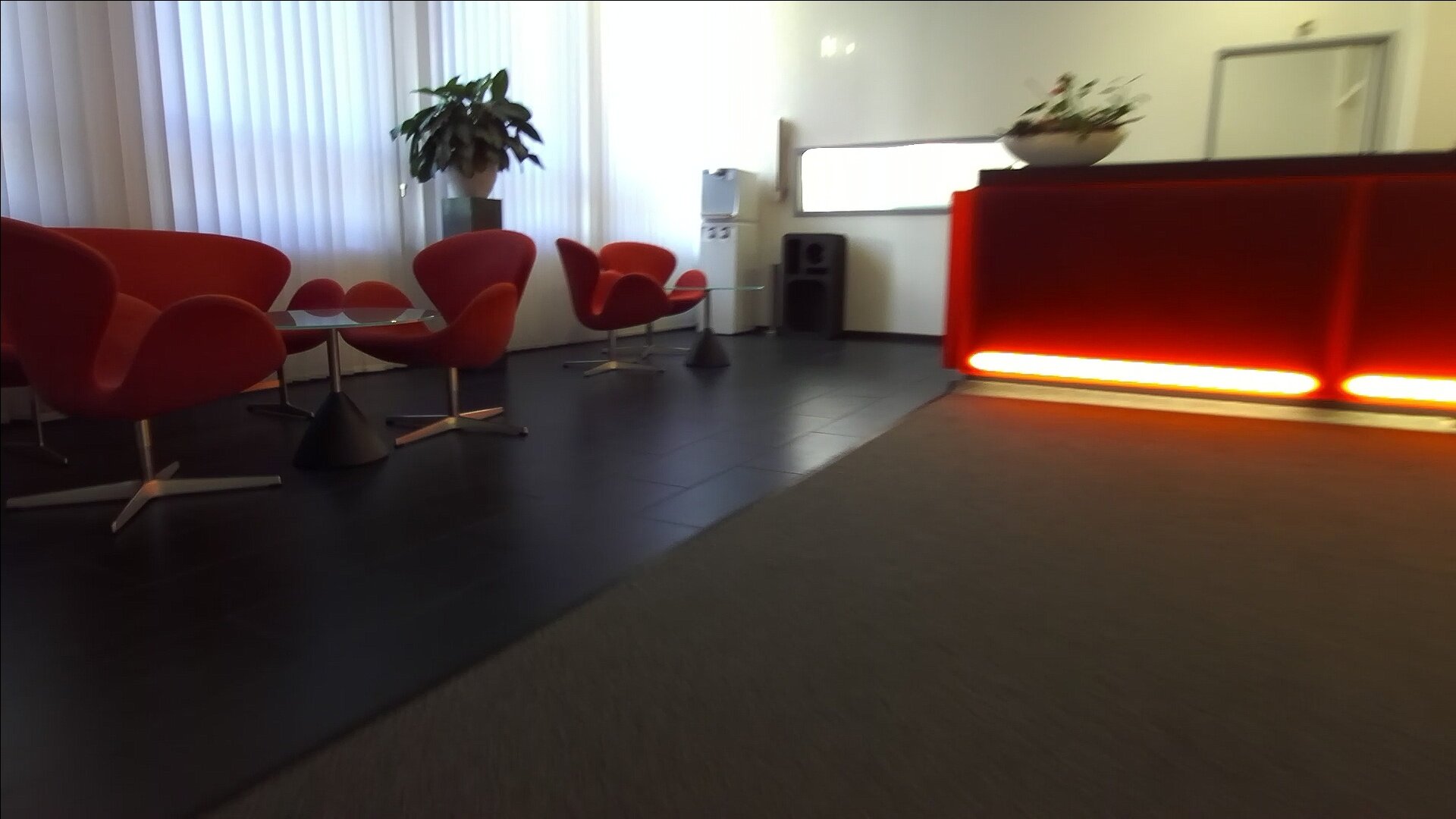}
    \includegraphics[width=0.1375\textwidth]{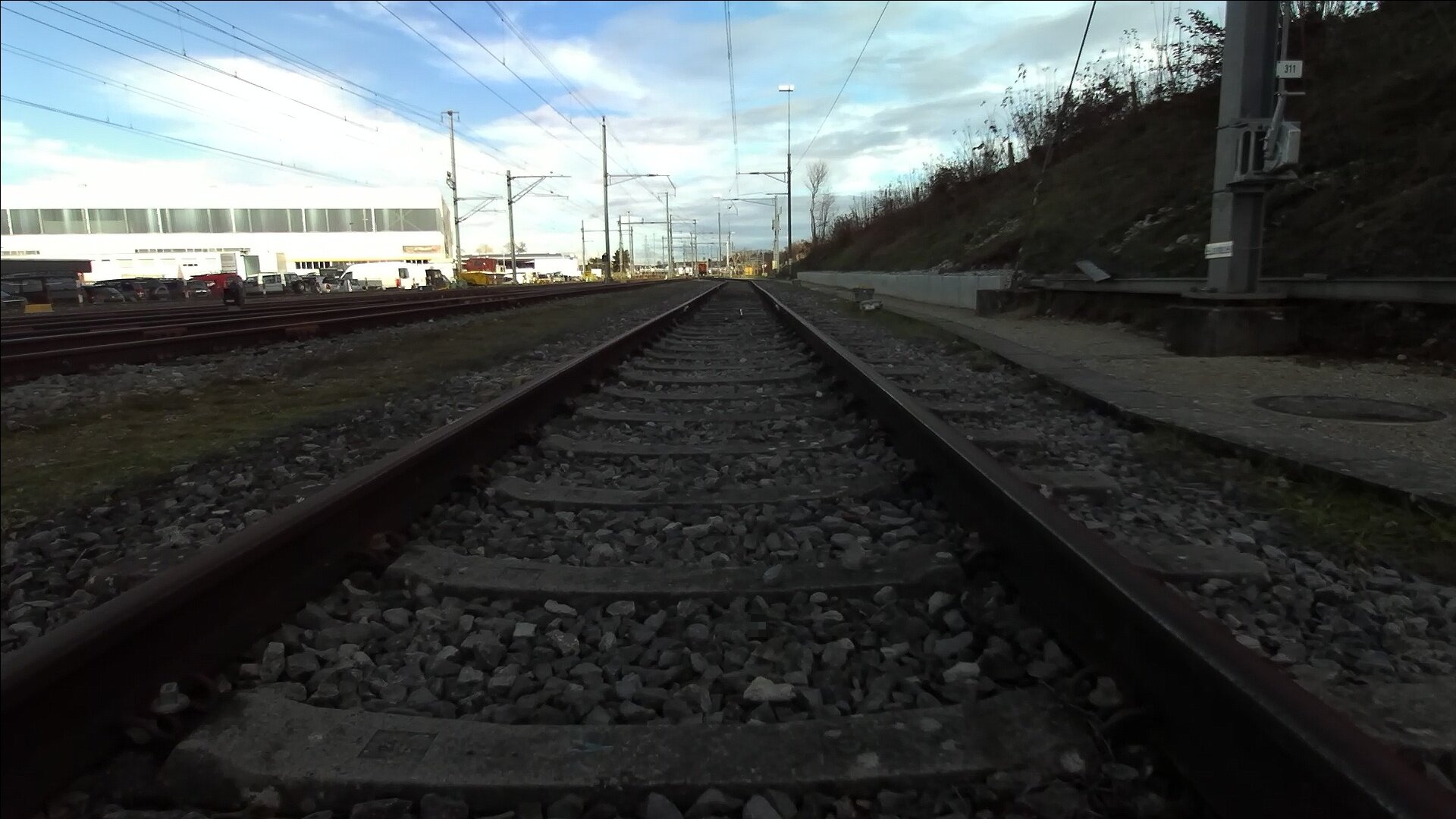}
    \caption{Example images from the selected missions reflecting diverse environments. From left to right: M10, M13, M19, M24, M34, M42, M44}
    \label{fig:mission_selection}

    \vspace{-1.0em}
\end{figure*}

\subsection{Sensor Selection}

The GrandTour dataset provides multi-sensor data via the Boxi multimodal payload architecture described in~\cite{frey2025boxi}. To isolate the specific hardware parameters required by the evaluated SLAM frameworks, we filter the available modalities to two camera systems, two IMUs, and one LiDAR sensor.

% For visual state estimation, we compare the Sevensense CoreResearch platform and the Stereolabs ZED2i camera system, excluding alternative payload cameras that lack native stereo or depth capabilities.
% The Sevensense suite provides global-shutter tracking, from which we utilize a front-facing RGB camera for monocular configurations and a hardware-synchronized, monochrome front-facing camera pair for stereo configurations. 
% These global-shutter streams record at 10~Hz with a 1440$\times$1080 resolution. 
% Conversely, the Stereolabs ZED2i camera utilizes front-facing rolling-shutter sensors operating at 15~Hz with a 1920$\times$1080 resolution, outputting synchronized RGB images and an active stereo depth map with a operational range of 15~m.

For visual estimation, we compare the global shutter Sevensense CoreResearch platform against the rolling shutter Stereolabs ZED2i camera system, excluding alternative cameras that lack native stereo or depth capabilities. 
The Sevensense suite records at 10~Hz with a 1440$\times$1080 resolution, providing a monocular RGB stream or a hardware synchronized monochrome stereo pair. 
Conversely, the ZED2i camera operates at 15~Hz with a 1920$\times$1080 resolution, outputting synchronized RGB images alongside an active stereo depth map with a 15~m operational range.

% For inertial navigation, we evaluate two time-synchronized IMUs representing distinct cost and performance brackets: a mid-range, industrial-grade Analog Devices ADIS15475-2 operating at 200~Hz and a high-end, tactical-grade Honeywell HG4930 operating at 100~Hz. 
% Other embedded inertial sensors integrated within the robot chassis or secondary receiver payloads are excluded due to unverified time-synchronization constraints or rapid dead-reckoning divergence noted in the Boxi payload documentation~\cite{frey2025boxi}.

For inertial navigation, we evaluate two time synchronized IMUs representing distinct cost and performance brackets. 
We utilize a mid range industrial grade Analog Devices ADIS15475-2 operating at 200~Hz and a high-end, tactical-grade Honeywell HG4930 operating at 100~Hz. 
Other embedded chassis or payload inertial sensors are excluded due to unverified synchronization constraints or rapid dead reckoning divergence noted in the baseline documentation~\cite{frey2025boxi}.

% For multi-modal fusion within the FAST-LIVO2 framework, the visual and inertial streams are paired with a fixed LiDAR sensor. Out of the payload options, the Livox Mid-360 is selected due to its tight spatial calibration and its superior 59$^\circ$ vertical field of view, which provides non-repetitive scanning up to a 40~m range with a ranging accuracy of 0.02~m. 
% Alternative scanning sensors, such as the chassis-integrated Velodyne or the payload-mounted Hesai units, are excluded because of severe structural payload occlusions or a restricted field-of-view.

For multimodal fusion within the FAST-LIVO2 framework, these streams are paired with a Livox Mid-360 LiDAR sensor. 
The Livox is selected due to its tight spatial calibration, non-repetitive scanning up to a 40~m range with 0.02~m accuracy, and a superior 59$^\circ$ vertical field of view. 
Alternative Velodyne or Hesai sensors are omitted due to severe structural payload occlusions or restricted fields of view.

\subsection{Mission Selection}

The GrandTour dataset contains a total of 49 missions. We select a representative subset of seven sequences based on ground-truth coverage, multi-modal sensor stream availability, and scene diversity. As detailed in Table~\ref{tab:mission_selection}, this selection spans indoor, urban, and unstructured outdoor landscapes, exposing the state estimation frameworks to a wide gradient of terrain topologies, textures, and illumination conditions.

Each sequence isolates distinct perceptual or mechanical stressors inherent to legged locomotion.
\textit{M10} spans an alpine region with deformable, low-texture snow and high-contrast sunlight.
\textit{M13} features a structured urban scene with sharp lighting transitions and stair-induced vertical mechanical excitation.
\textit{M19} introduces violent motion on rough mountain trails, including a 180$^\circ$ reverse loop where sudden orientation swaps degrade visuals and challenge place recognition.
For loop-free validation, \textit{M24} covers a muddy construction site with highly compliant ground, wet grass, and standing water.
\textit{M34} evaluates severe perceptual shifts during an outdoor-to-tunnel transition, where continuous turns trigger camera auto-exposure lag.
\textit{M42} introduces dynamic visual obstructions from pedestrians and vehicles during an urban outdoor-to-indoor transition.
Finally, \textit{M44} evaluates long, linear tracking along a railway over loose soil and gravel, where repetitive structures cause visual aliasing before a final, visually degraded reverse loop.

% Each selected sequence isolates specific perceptual or mechanical stress cases inherent to legged locomotion. 
% \textit{M10} covers an alpine region with deformable, snow-covered terrain characterized by low visual texture and high-contrast sunlight. 
% \textit{M13} represents a structured urban scene with sharp lighting transitions and a stair traversal sequence that induces severe vertical mechanical excitation.
% \textit{M19} introduces a highly violent motion profile due to rough mountain trails, including a 180$^\circ$ reverse-pass loop where the sudden orientation swap causes severe visual degradation that challenges place recognition.
% For loop-free validation, \textit{M24} provides a trajectory across an industrial construction site with highly compliant ground surfaces like wet grass and standing water. 
% Severe perceptual shifts are evaluated in \textit{M34}, where the robot transitions from an outdoor environment into an illuminated underground tunnel network requiring continuous turning maneuvers that trigger camera auto-exposure lag. 
% \textit{M42} introduces dynamic visual obstructions via moving pedestrians and vehicles during a mixed urban outdoor-to-indoor transition. 
% Finally, \textit{M44} evaluates a long, linear tracking sequence along a railway track over loose soil and gravel, featuring repetitive geometric structures that cause visual aliasing, concluded by a visually degraded reverse-pass loop.

\subsection{Evaluation Protocol}

We quantitatively evaluate the state estimation frameworks across three primary dimensions: localization accuracy, tracking robustness, and computational resource utilization. 
Following \cite{sturm2012tumrgbd}, localization accuracy is quantified using the Root-Mean-Square Error (RMSE) of the Absolute Trajectory Error (ATE) and the Relative Pose Error (RPE), both computed via the open-source \texttt{evo} evaluation toolkit~\cite{grupp2017evo}. 
The ATE evaluates global consistency by assessing the absolute positional deviation between the estimated trajectory and the ground truth after spatial alignment. 
Following the benchmarking protocols defined in~\cite{zhang2018tutorial}, this spatial synchronization is executed using Umeyama alignment~\cite{umeyama1991alignment}. 
For trajectories generated by monocular configurations, the alignment backend includes scale optimization via a similarity transformation in $Sim(3)$ to account for the inherent scale ambiguity of single-camera pipelines. 
Conversely, for stereo, RGB-D, and multi-modal configurations, scaling is disabled to enforce a strict rigid body transformation in $SE(3)$. 
The RPE measures local accuracy by evaluating the relative transformation errors between consecutive poses over short intervals. 
For both metrics, only the translational components are considered because the reference ground-truth system lacks reliable orientation information~\cite{frey2026grandtour}. 
% Evaluating only translation remains highly representative, as unmodeled rotational drift naturally compounds over time and manifests as distinct translational offsets in the later stages of the trajectory.

To account for algorithmic non-determinism, we repeat each experimental configuration three times on every sequence, and report the average RMSE metrics.
Tracking robustness is enforced via a strict scenario duration coverage threshold. 
% Specifically, any tracking run that fails to maintain valid pose estimates for at least 80\% of the total mission duration is classified as a failure, similar to the ROVER benchmark evaluation protocol~\cite{schmidt2025rover}. 
Following the evaluation protocol of the ROVER benchmark~\cite{schmidt2025rover}, a run is classified as a failure if valid pose estimates are available for less than 80\% of the mission duration.
These invalid trajectories are excluded from the accuracy mean computations to prevent partial or aborted tracking runs from artificially lowering the reported metrics.

Computational resource footprints are benchmarked by continuously logging the CPU utilization and RAM consumption of the active ROS nodes. 
To ensure a standardized and fair execution environment, all experiments are conducted on an AMD
Ryzen 9 5900X 12-Core processor with 64~GB of RAM, operating within an Ubuntu 20.04 Docker container configured with ROS Noetic except DPV-SLAM, which was evaluated in an Ubuntu 22.04 Docker container.

%%%%%%%%%%%%%%%%%%%%%%%%%%%%%%%%%%%%%%%%%%%%%%%%%%%%%%%%%%%%%%%%%%%%%%%%%%%%%%%%
\section{Results and Discussion}
\label{sec:eval}

We systematically evaluate SLAM performance across four core dimensions to isolate how sensor configuration choices mitigate or exacerbate quadrupedal motion artifacts.
First, we analyze visual configurations across monocular, stereo, and RGB-D setups. 
Second, we assess global versus rolling shutter technologies. 
Third, we evaluate the impact of inertial data integration across multiple hardware tiers. 
Finally, we conduct a cross architectural comparison that benchmarks the optimal configuration of each framework against onboard computational resource utilization.

\subsection{Visual Configurations}

\begin{table*}
\centering
\caption{Evaluation of ORB-SLAM3 and RTAB-Map configurations using RMSE ATE and RMSE RPE. Fields marked with "-" had no successful run for the specified mission and used sensor configuration. Results marked with (*) were calculated using only the successful runs. Best method-specific results highlighted as \colorbox{colorFst}{\textbf{first}}, \colorbox{colorSnd}{second}, \colorbox{colorTrd}{third}.}
\label{tab:mono_stereo_rgbd}
\small % Scales font slightly down to prevent text crowding
\setlength{\tabcolsep}{3pt} % Tighter padding between columns to fit all metrics cleanly
\footnotesize
\begin{tabularx}{\linewidth}{l *{8}{Y} c *{8}{Y}}
\toprule
\multirow{2}{*}{\textbf{Config}} & \multicolumn{8}{c}{\textbf{RMSE ATE [m] ($\downarrow$)}} & & \multicolumn{8}{c}{\textbf{RMSE RPE [m] ($\downarrow$)}} \\
\cmidrule{2-9} \cmidrule{11-18}
 & M10 & M13 & M19 & M24 & M34 & M42 & M44 & Mean & & M10 & M13 & M19 & M24 & M34 & M42 & M44 & Mean \\
\midrule

\multicolumn{18}{l}{\textbf{ORB-SLAM3}} \\
\quad Mono, Sevensense   & \rd 0.357 & \nd 0.297 & 20.204     & \fst 0.136 & \nd 16.794 & \nd 0.268  & \nd 11.937 & \rd 6.586  & & \rd 0.032 & \nd 0.052 & 3.768      & \nd 0.020  & \nd 2.316  & \nd 0.057  & \nd 1.180  & \rd 1.010 \\
\quad Stereo, Sevensense & \fst 0.052 & \fst 0.124  & \fst 0.255 & \nd 0.137  & \fst 0.866 & \fst 0.233 & \fst 0.641 & \fst 0.330 & & \fst 0.005 & \fst 0.017  & \fst 0.011 & \fst 0.013 & \fst 0.022 & \fst 0.037 & \fst 0.017 & \fst 0.017 \\
\quad Mono, ZED2i        & 1.128      & 8.585      & 15.558     & 3.163      & --         & --         & \rd 25.128 & 10.712* & & 0.099      & 2.429      & 2.480      & 0.426      & --         & --         & \rd 3.627  & 1.812* \\
\quad Stereo, ZED2i      & \nd 0.181  & \rd 0.437  & \nd 0.795  & \rd 0.339  & 28.991     & 10.913     & 30.715     & \nd 6.494  & & \nd 0.020  & \rd 0.053  & \nd 0.038  & \rd 0.039  & \rd 4.647  & \rd 1.785  & 3.980      & \nd 0.826 \\
\quad RGB-D, ZED2i       & 2.646      & 5.499      & \rd 1.468  & 0.865      & \rd 26.175 & \rd 10.390 & 30.595     & 8.214      & & 0.572      & 1.531      & \rd 0.055  & 0.089      & 4.897      & 1.826      & 3.880      & 1.225 \\

\midrule
\multicolumn{18}{l}{\textbf{RTAB-Map}} \\
\quad Stereo, Sevensense & \fst 0.207 & \fst 0.375 & \fst 0.294 & \fst 0.434 & -- & \fst 2.869 & \fst 1.586 & \fst 0.579* & & \nd 0.048 & \nd 0.107 & \fst 0.034 & \fst 0.060 & -- & \fst 0.333 & \fst 0.039 & \fst 0.058* \\
\quad Stereo, ZED2i      & \nd 0.342  & \nd 0.574  & \nd 1.009  & \rd 1.999  & -- & --         & \rd 8.884  & \rd 2.562* & & \fst 0.046 & \fst 0.098 & \nd 0.039  & \rd 0.298  & -- & --         & \rd 0.361  & \rd 0.168* \\
\quad RGB-D, ZED2i       & \rd 1.552  & \rd 3.045  & \rd 1.891  & \nd 1.687  & -- & --         & \nd 3.892  & \nd 2.413* & & \rd 0.133  & \rd 0.226  & \rd 0.083  & \nd 0.144  & -- & --         & \nd 0.116  & \nd 0.140* \\
\bottomrule
\end{tabularx}

\vspace{-1.0em}

\end{table*}

%%%%%%%%%%%%%%%%%%%%%%%%%%%%%%%%%%%%%%%%%%%%%%
We evaluate monocular, stereo, and RGB-D sensor configurations for ORB-SLAM3 alongside stereo and RGB-D baselines for RTAB-Map across seven distinct missions, as summarized in Table \ref{tab:mono_stereo_rgbd}. 
Cross referencing these results with the environmental attributes detailed in Table \ref{tab:mission_selection} provides a clear physical context for the performance variations observed across different terrain sequences.

Within the ORB-SLAM3 framework, the configurations are divided between the Sevensense CoreResearch and Stereolabs ZED2i camera hardware suites. 
Within the Sevensense system, the stereo configuration outperforms the monocular baseline across nearly all evaluation runs. 
A minor exception occurs during mission \textit{M24} where the monocular configuration achieves a marginally lower ATE than the stereo baseline, though the stereo mode retains superior relative pose accuracy. 
When utilizing the ZED2i camera, the RGB-D arrangement displays improved tracking accuracy over the monocular setup, yet the corresponding stereo configuration ultimately captures the highest overall precision. 
Extrapolating across all tested variations, the stereo configuration using the Sevensense platform yields the most accurate mean performance profile for ORB-SLAM3. 
Conversely, the monocular configurations experience severe scale drift during missions \textit{M19} on low-texture alpine terrain, \textit{M34} during dark underground transitions, and \textit{M44} along outdoor industrial railway tracks. 
Parallel trends are observable with RTAB-Map where the Sevensense hardware operating in a stereo configuration similarly delivers the highest overall tracking accuracy. However, when utilizing the ZED2i camera suite, the RTAB-Map framework diverges slightly from ORB-SLAM3 as its RGB-D configuration yields lower errors than the corresponding stereo arrangement.

\begin{figure}[tbp]
    \centering
    % Top Subfigure: Rolling Shutter
    \begin{subfigure}{\columnwidth}
        \centering
        \caption*{ZED2i - Rolling Shutter $\mid$ Matches: 33}
        \includegraphics[width=\textwidth]{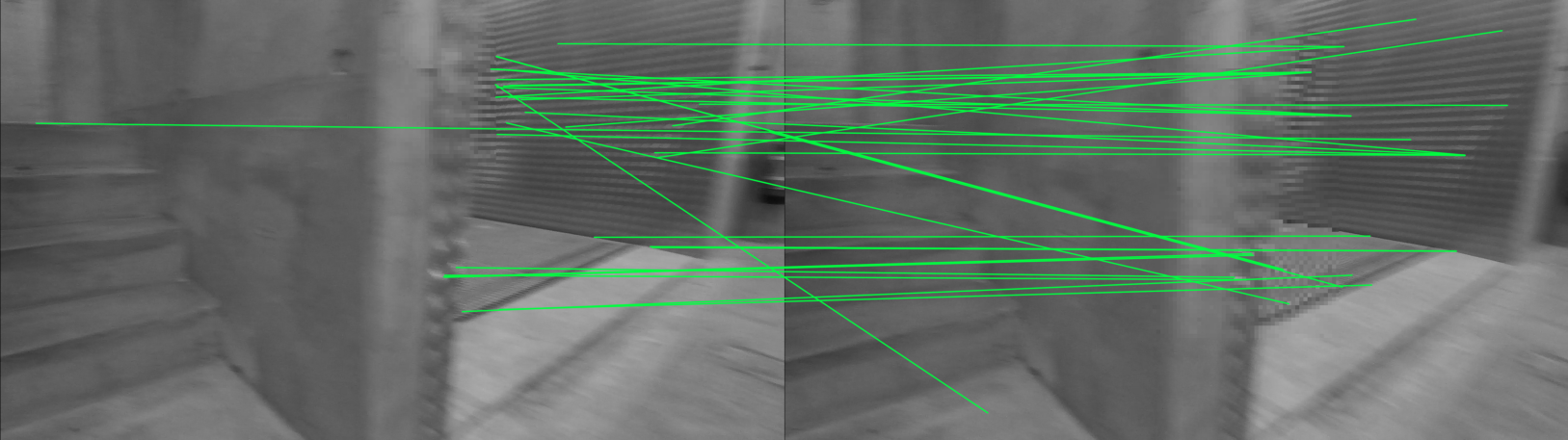}
        \label{fig:rolling_shutter}
    \end{subfigure}
    
    \vspace{-1.5em}
    
    % Bottom Subfigure: Global Shutter
    \begin{subfigure}{\columnwidth}
        \centering
        \caption*{Sevensense - Global Shutter $\mid$ Matches: 188}
        \includegraphics[width=\textwidth]{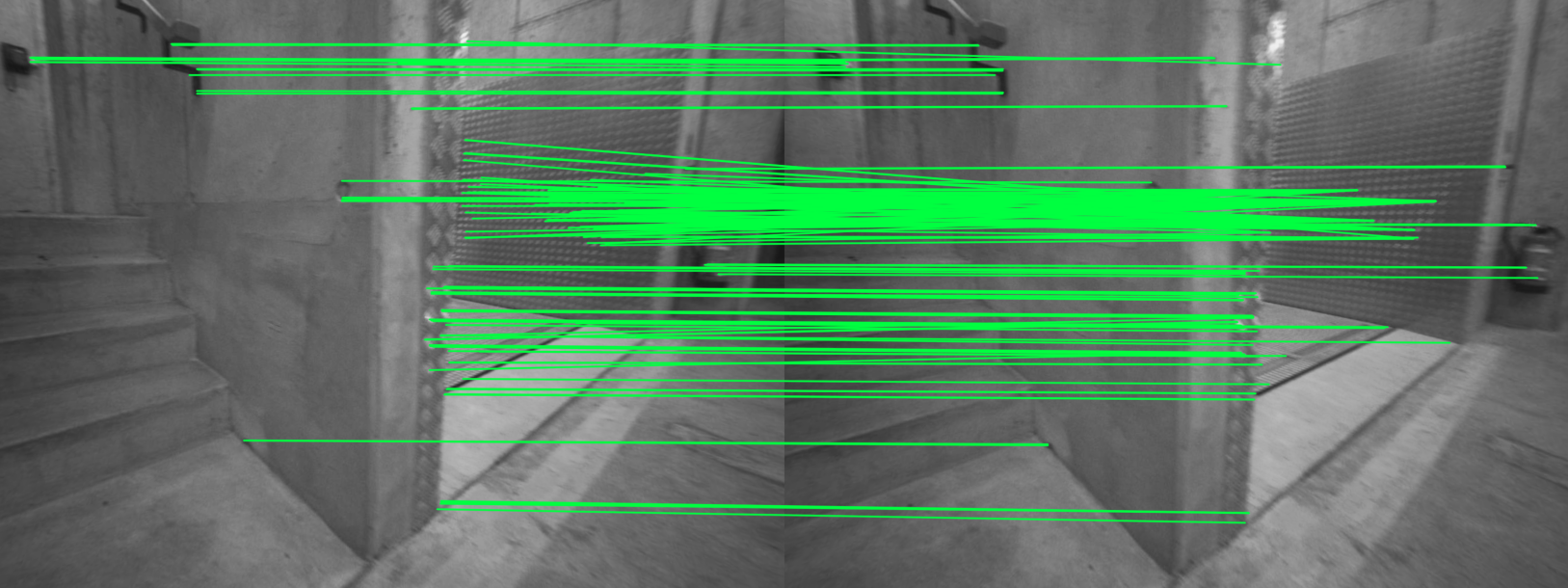}
        \label{fig:global_shutter}
    \end{subfigure}

    \vspace{-0.75em}

    \caption{ORB feature matching during high-velocity rotation (\textit{M34}). The global shutter (bottom) maintains robust tracking with 188 consistent matches, whereas the rolling shutter (top) drops to 33 matches due to motion blur, introducing incorrect correspondences that degrade pose estimation.}
    \label{fig:shutter_comparison}

    \vspace{-0.75em}
\end{figure}

Several configurations suffer complete tracking failures under the demanding operational limits of missions \textit{M34} and \textit{M42} which feature dark subterranean segments and dynamic visual obstructions. 
The RTAB-Map stereo configuration using the Sevensense camera fails to complete mission \textit{M34}, whereas both the stereo and RGB-D configurations utilizing the ZED2i camera fail during both sequences. 
For ORB-SLAM3, terminal tracking failures are restricted to the monocular ZED2i configuration during these two specific challenges.
The failure analysis indicates that ZED2i configurations diverge during transitions from stationary postures to rapid turning maneuvers over textureless surfaces, preventing frame-to-frame feature association.
As shown in Fig. \ref{fig:shutter_comparison} (top), motion blur severely degrades visual tracking, causing a drop in matched correspondences.
A matching failure under identical geometric conditions causes the Sevensense variant of RTAB-Map to fail during the underground turns of mission \textit{M34}.
Additionally, the monocular ZED2i variant of ORB-SLAM3 loses features during indoor turns and fails to maintain trajectory consistency during the reverse pass loop of \textit{M42}.
These tracking anomalies expose an inherent vulnerability in both feature-based frameworks when subjected to aggressive rotational velocities with minimal translational displacement.

Architecturally, performance ranking of visual configurations under quadruped locomotion mirrors wheeled platforms like the ROVER benchmark~\cite{schmidt2025rover}. 
Both domains suffer severe monocular drift, which is heavily mitigated by anchoring tracking with absolute geometric scale via stereo or RGB-D sensors. 
Additionally, feature-based tracking during aggressive rotational maneuvers with minimal translation remains a shared algorithmic bottleneck across both platform classes.

%%%%%%%%%%%%%%%%%%%%%%%%%%%%%%%%%%%%%%%%%%%%%%

\subsection{Shutter Technologies}
\begin{table*}
\centering
\caption{Comparative Evaluation of SLAM Configurations (Global vs. Rolling Shutter) using RMSE ATE and RMSE RPE. Global shutter configurations consistently yield significantly lower mean tracking errors and superior trajectory robustness compared to rolling shutter variants.}
\label{tab:global_rolling}
\small % Scaled down slightly to comfortably fit all 18 columns across the page
\setlength{\tabcolsep}{2.5pt} % Optimizes padding between data intersections
\footnotesize
\begin{tabularx}{\linewidth}{l l *{8}{Y} c *{8}{Y}}
\toprule
\multicolumn{2}{c}{\textbf{Config}} & \multicolumn{8}{c}{\textbf{RMSE ATE [m] ($\downarrow$)}} & & \multicolumn{8}{c}{\textbf{RMSE RPE [m] ($\downarrow$)}} \\
\cmidrule{3-10} \cmidrule{12-19}
& & M10 & M13 & M19 & M24 & M34 & M42 & M44 & Mean & & M10 & M13 & M19 & M24 & M34 & M42 & M44 & Mean \\
\midrule

\multicolumn{19}{l}{\textbf{ORB-SLAM3, mono}} \\
& Global  & 0.357 & \rd 0.297 & 20.204 & \rd 0.136 & 16.794 & 0.268 & 11.937 & 6.586 & & 0.032 & \rd 0.052 & 3.768 & \rd 0.020 & 2.316 & \rd 0.057 & 1.180 & 1.010 \\
& Rolling & 1.128 & 8.585 & 15.558 & 3.163 & -- & -- & 25.128 & 10,712* & & 0.099 & 2.429 & 2.480 & 0.426 & -- & -- & 3.627 & 1.812* \\
\midrule

\multicolumn{19}{l}{\textbf{ORB-SLAM3, stereo}} \\
& Global  & \fst 0.052 & \nd 0.124 & \nd 0.255 & 0.137 & \nd 0.866  & \rd 0.233  & \rd 0.641  & \fst 0.330  & & \fst 0.005 & \fst 0.017 & \fst 0.011 & \fst 0.013 & \fst 0.022 & \fst 0.037 & \fst 0.017 & \fst 0.017 \\
& Rolling &  0.181 & 0.437 & 0.795 & 0.339 & 28.991 & 10.913 & 30.715 & 10.339 & & \nd 0.020 & 0.053 & \rd 0.038 & 0.039 & 4.647 & 1.785 & 3.980 & 1.509 \\
\midrule

\multicolumn{19}{l}{\textbf{RTAB-Map, stereo}} \\
& Global  & 0.207 & 0.375 & \rd 0.294 & 0.434 & --    & 2.869 & 1.586 & 0.579* & & 0.048 & 0.107 & \nd 0.034 & 0.060 & --    & 0.333 & \rd 0.039 & 0.058* \\
& Rolling & 0.342 & 0.574 & 1.009 & 1.999 & -- & -- & 8.884 & 2.562* & & 0.046 & 0.098 & 0.039 & 0.298 & -- & -- & 0.361 & 0.168* \\
\midrule

\multicolumn{19}{l}{\textbf{FAST-LIVO2, livox, ADIS}} \\
& Global  & \nd 0.058 & \fst 0.044 & 1.261 & \fst 0.069 & \rd 1.662  & \fst 0.057 & \nd 0.318 & \nd 0.495 & & \rd 0.025 & \nd 0.029 & 0.139 & \nd 0.019 & \rd 0.181 & \nd 0.046 & \nd 0.033 & \rd 0.067 \\
& Rolling & \rd 0.061 & \rd 0.151 & \fst 0.161 & \nd 0.070 & 19.709 & \nd 0.061 & \fst 0.303 & 2.931 & & 0.034 & 0.097 & 0.058 & 0.033 & 3.286 & \nd 0.046 & 0.070 & 0.517 \\
\midrule

\multicolumn{19}{l}{\textbf{DPV-SLAM}} \\
& Global  & 0.400 & 0.479 & 0.784 & 0.977 & \fst 0.812  & 0.490  & 2.825  & \rd 0.967  & & 0.032 & 0.068 & 0.061 & 0.096 & \nd 0.045 & 0.081 & 0.053 & \nd 0.062 \\
& Rolling & 5.284 & 11.716 & 8.425 & 6.529 & 17.150 & 5.113  & 27.619 & 12.076 & & 0.549 & 2.088 & 2.483 & 0.894 & 2.011 & 0.967 & 1.535 & 1.504 \\
\midrule
\midrule

\multicolumn{19}{l}{\textbf{Mean}} \\
& Global  & 0.215 & 1.012 & 4.560 & 0.350 & 5.033  & 0.783 & 3.461 & 1.898 & & 0.028 & 0.250 & 0.803 & 0.042 & 0.641 & 0.111 & 0.264 & 0.271 \\
& Rolling & 1.399 & 4.293 & 5.189 & 2.420 & 22.849 & 5.362 & 18.530 & 7.028 & & 0.150 & 0.953 & 1.019 & 0.338 & 3.314 & 0.933 & 1.914 & 1.102 \\
\bottomrule
\end{tabularx}

\end{table*}

% TODO: explain outliers
% In addition to the monocular, stereo, and RGB-D configurations, we evaluate global and rolling shutter sensor configurations for RTAB-Map, ORB-SLAM3 (monocular and stereo variants), FAST-LIVO2 (utilizing the Livox LiDAR and ADIS IMU), and DPV-SLAM. The results are compiled in Table \ref{tab:global_rolling}.

We evaluate global and rolling shutter configurations for RTAB-Map, ORB-SLAM3 monocular and stereo variants, FAST-LIVO2 operating with the Livox LiDAR and ADIS IMU, and DPV-SLAM. 
The quantitative performance metrics across configurations are shown in Table \ref{tab:global_rolling}.

For every SLAM framework tested, the global shutter configuration achieves the lowest mean ATE and RPE. 
Specifically, the ORB-SLAM3 stereo configuration with global shutter exposure captures the highest overall precision, securing the lowest RPE across all evaluated missions alongside the best mean ATE. 
Because rolling shutter sensors read out image rows sequentially over a finite exposure interval, aggressive angular velocities from the quadruped introduce severe non-rigid geometric distortions. 
This sampling artifact directly causes optimization divergence in demanding environments. 
For instance, while the global shutter variant of RTAB-Map encounters tracking failure only during the dark subterranean turns of mission \textit{M34}, the rolling shutter configurations for both stereo RTAB-Map and monocular ORB-SLAM3 suffer catastrophic tracking divergence across both missions \textit{M34} and \textit{M42}.

\begin{figure}
    \centering
    \includegraphics[width=1.\columnwidth]{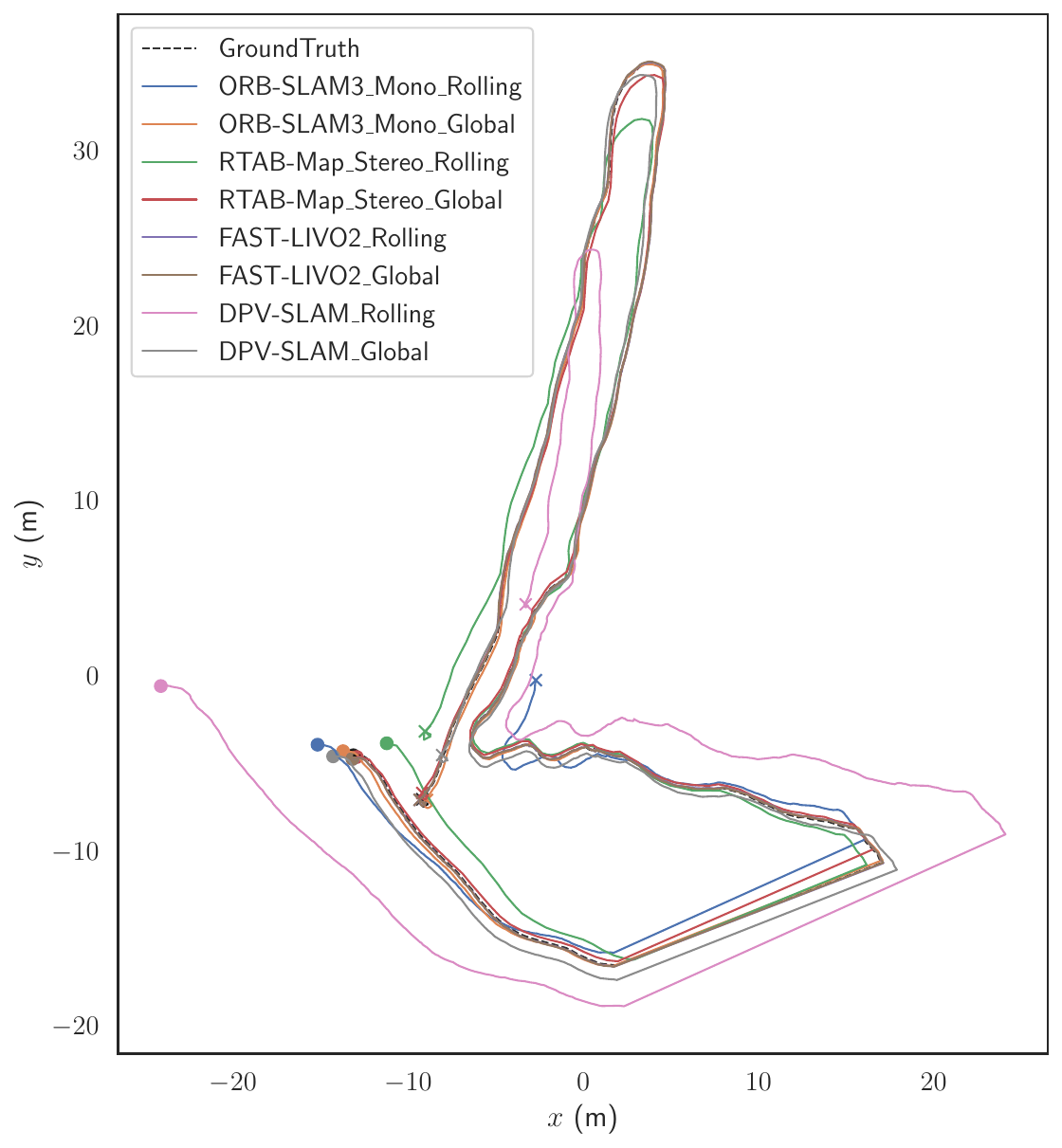}
    \caption{Estimated trajectories for mission \textit{M24} illustrating that global shutter prevents terminal tracking failures during rapid turning maneuvers and reduces drift across all evaluated SLAM frameworks.}
    \label{fig:global_rolling}

    \vspace{-0.75em}
\end{figure}

Qualitative trajectory comparisons for global and rolling shutter configurations are illustrated in Fig. \ref{fig:global_rolling}. 
The paths demonstrate that global shutter consistently improves tracking accuracy across all frameworks by eliminating motion artifacts where rolling shutter distortions heavily degrade visual feature tracking. 
% This accuracy gap is highly pronounced for ORB-SLAM3 stereo, FAST-LIVO2, DPV-SLAM and RTAB-Map where rolling shutter distortions heavily degrade visual feature tracking. 
Furthermore, during this specific evaluation sequence, the rolling shutter configuration of ORB-SLAM3 experiences terminal tracking failure during rapid turning maneuvers whereas the corresponding global shutter configuration successfully recovers the entire trajectory, see Fig. \ref{fig:shutter_comparison}. This outcome shows that global shutter technology can mitigate embodiment-induced visual degradation and improve accuracy and robustness in visual feature tracking, which is fundamental for legged robots.

% \begin{figure}[!htbp]
%     \centering
%     \includegraphics[width=0.42\columnwidth]{1732543285.516582_alph_dist.png}
%     \hfill
%     \includegraphics[width=0.56\columnwidth]{1732543285.517922_zed2i.png}
%     \caption{Images captured at the same time in a fast turn of mission LEICA-1: the left image was captured with the Sevensense global shutter camera, while the right image was captured with the ZED2i camera. For the ZED2i, rolling-shutter distortion and motion blur can negatively impact visual feature tracking.}
%     \label{fig:zed2i_vs_alphasense}
% \end{figure}

%%%%%%%%%%%%%%%%%%%%%%%%%%%%%%%%%%%%%%%%%%%%%%%%%%%%%%%%%%%%%%%%%%%%%%%%%%%%
\subsection{Inertial Data Integration}
\begin{table*}
\centering
\caption{ Comparative Evaluation of Inertial SLAM Configurations (ADIS vs. Honeywell IMU) using RMSE ATE and RMSE RPE. The high-end Honeywell IMU improves mean ATE and RPE for FAST-LIVO2, while inertial integration can reduce robustness in vision-dominant methods compared to purely visual configurations.}
\label{tab:imu}
\small % Scaled down slightly to comfortably fit all 18 columns across the page
\setlength{\tabcolsep}{2.5pt} % Optimizes padding between data intersections
\footnotesize
\begin{tabularx}{\linewidth}{l l *{8}{Y} c *{8}{Y}}
\toprule
\multicolumn{2}{c}{\textbf{Config}} & \multicolumn{8}{c}{\textbf{RMSE ATE [m] ($\downarrow$)}} & & \multicolumn{8}{c}{\textbf{RMSE RPE [m] ($\downarrow$)}} \\
\cmidrule{3-10} \cmidrule{12-19}
 & & M10 & M13 & M19 & M24 & M34 & M42 & M44 & Mean & & M10 & M13 & M19 & M24 & M34 & M42 & M44 & Mean \\
\midrule

\multicolumn{19}{l}{\textbf{ORB-SLAM3, mono-inertial}} \\
& ADIS      & 0.978 & 2.922 & 2.704 & 1.440 & 6.711 & --    & \rd 1.884 & 2.773* & & 0.110 & 0.227 & 0.193 & 0.136 & 0.526 & --    & 0.050 & 0.207* \\
& Honeywell & 0.184 & 1.072 & \rd 0.286 & 0.181 & 0.926 & --    & 2.496 & 0.858* & & \rd 0.020 & 0.054 & \rd 0.019 & 0.027 & \rd 0.066 & --    & 0.052 & 0.040* \\
\midrule

\multicolumn{19}{l}{\textbf{ORB-SLAM3, stereo-inertial}} \\
& ADIS      & 0.103 & \rd 0.260 & \nd 0.249 & \rd 0.133 & \rd 0.360 & 0.290 & --    & 0.229* & & \nd 0.009 & \fst 0.010 & \fst 0.007 & \fst 0.014 & \fst 0.015 & \fst 0.023 & --    & 0.014* \\
& Honeywell & \rd 0.095 & \nd 0.147 & --    & 0.135 & \nd 0.236 & \rd 0.229 & \nd 0.516 & 0.168* & & \fst 0.008 & \nd 0.011 & --    & \nd 0.015 & \fst 0.015 & \nd 0.036 & \fst 0.013 & 0.017* \\
\midrule

\multicolumn{19}{l}{\textbf{RTAB-Map, stereo-inertial}} \\
& ADIS      & 0.243 & 0.429 & 0.424 & 0.632 & --    & 2.933 & 1.927 & 1.098* & & 0.032 & 0.091 & \nd 0.012 & 0.046 & --    & 0.335 & \rd 0.030 & 0.091* \\
& Honeywell & 0.219 & 0.442 & 0.558 & 0.957 & --    & 2.920 & 2.376 & 1.246* & & 0.029 & 0.092 & \nd 0.012 & 0.046 & --    & 0.330 & \nd 0.028 & 0.089* \\
\midrule

\multicolumn{19}{l}{\textbf{FAST-LIVO2, livox, sevensense}} \\
& ADIS      & \fst 0.058 & \fst 0.044 & 1.261 & \fst 0.069 & 1.662 & \nd 0.057 & \fst 0.318 & \nd 0.495 & & 0.025 & 0.029 & 0.139 & \rd 0.019 & 0.181 & 0.046 & 0.033 & \nd 0.067 \\
& Honeywell & \nd 0.059 & \fst 0.044 & \fst 0.126 & \nd 0.071 & \fst 0.063 & \fst 0.053 & \fst 0.318 & \fst 0.105 & & 0.022 & \rd 0.028 & 0.052 & 0.022 & \nd 0.027 & \rd 0.037 & 0.060 & \fst 0.035 \\
\midrule
\midrule

\multicolumn{19}{l}{\textbf{Mean}} \\
& ADIS      & 0.346 & 0.914 & 1.159 & 0.568 & 2.911 & 1.094 & 1.376 & 1.149 & & 0.044 & 0.089 & 0.088 & 0.054 & 0.241 & 0.135 & 0.038 & 0.095 \\
& Honeywell & 0.139 & 0.426 & 0.324 & 0.336 & 0.409 & 1.067 & 1.426 & 0.594 & & 0.020 & 0.046 & 0.028 & 0.028 & 0.036 & 0.134 & 0.038 & 0.045 \\
\bottomrule
\end{tabularx}

\vspace{-1.0em}

\end{table*}

% TODO: explain outliers

Table \ref{tab:imu} outlines the tracking metrics obtained across different IMU sensor categories, comparing a mid-range ADIS IMU against a high-end Honeywell IMU. 
To isolate the specific influence of inertial data integration, the global shutter Sevensense camera platform is maintained for all visual-inertial configurations due to its superior baseline performance, alongside the active Livox LiDAR for the FAST-LIVO2 framework. 

Utilizing the high-end Honeywell IMU improves the ATE across a majority of the tested configurations. 
The mid-range ADIS sensor only yields lower ATE for the stereo-inertial RTAB-Map configuration, while capturing superior RPE within the stereo-inertial variant of ORB-SLAM3.
Inertial integration fails to improve the tracking resilience of RTAB-Map.
Across both sensors, RTAB-Map suffers tracking divergence during the exact same textureless turning sequences that caused failures in the pure visual baseline, demonstrating that inertial data cannot compensate for severe frame-to-frame feature degradation. 
Furthermore, IMU integration actively degrades the mean ATE and RPE metrics for RTAB-Map. 
For ORB-SLAM3, the inertial modes exhibit reduced operational robustness across missions compared to the pure visual global shutter results, wehre each mission was successful. 
The mono-inertial configuration experiences complete tracking failure during the rapid rotational maneuvers of mission \textit{M42}. 
Additionally, both inertial variants exhibit tracking anomalies during the initial reverse pass loop sequences. 
Notably, the stereo-inertial configuration encounters catastrophic failures during mission \textit{M19} when paired with the Honeywell IMU and during mission \textit{M44} when utilizing the ADIS IMU. 
During these specific sequences, ORB-SLAM3 terminates abruptly without generating distinct error log messages. 
Because the corresponding pure visual configurations successfully track these exact trajectories, we hypothesize that these terminal failures are directly induced by the inertial integration backend, potentially due to numerical instabilities or state propagation divergence triggered by the high-frequency mechanical shocks of quadrupedal locomotion. 
Conversely, for the filter-based FAST-LIVO2 framework, utilizing the high-end Honeywell IMU significantly suppresses estimator drift, outperforming the best optimization-based configurations in mean ATE.

This performance degradation highlights a major architectural divergence from wheeled platforms like the ROVER benchmark~\cite{schmidt2025rover}. 
While wheeled rover evaluations demonstrate that coupling inertial measurements with visual frontends consistently stabilizes state estimation and reduces drift, this classic synergy fails under legged locomotion. 
The intense high frequency mechanical shocks and intermittent foot contact impacts unique to quadrupeds inject severe noise into tightly coupled optimization backends, rendering visual-inertial integration a liability for vision-centric frameworks like ORB-SLAM3 and RTAB-Map rather than an asset.

% Qualitative trajectory paths across the different inertial configurations are illustrated in Figure \ref{fig:imu}. 
% The paths demonstrate that upgrading from the mid-range ADIS IMU to the high-end Honeywell IMU consistently minimizes tracking drift for both ORB-SLAM3 and FAST-LIVO2. 
% Additionally, the qualitative plots show that RTAB-Map achieves a tighter trajectory alignment when coupled with the Honeywell sensor compared to the ADIS sensor, which accurately reflects the quantitative performance trends documented in Table \ref{tab:imu}.

% \TODO{reference method selection characterization and why specific methods worked better/had better results (robustness, accuracy, resource usage)}

% \begin{figure}[h]
%     \centering
%     \includegraphics[width=.555\columnwidth]{imu.png}
%     % \caption{Estimated trajectories using different IMUs for mission 24}
%     \caption{Estimated trajectories for mission \textit{M24} demonstrating that upgrading from a mid-range ADIS IMU to a high-end Honeywell IMU consistently suppresses tracking drift and reduces spatial errors across frameworks.}
%     \label{fig:imu}
% \end{figure}

\subsection{Cross Architectural Comparison}

\begin{figure}[htb]
    \centering
    \includegraphics[width=\columnwidth]{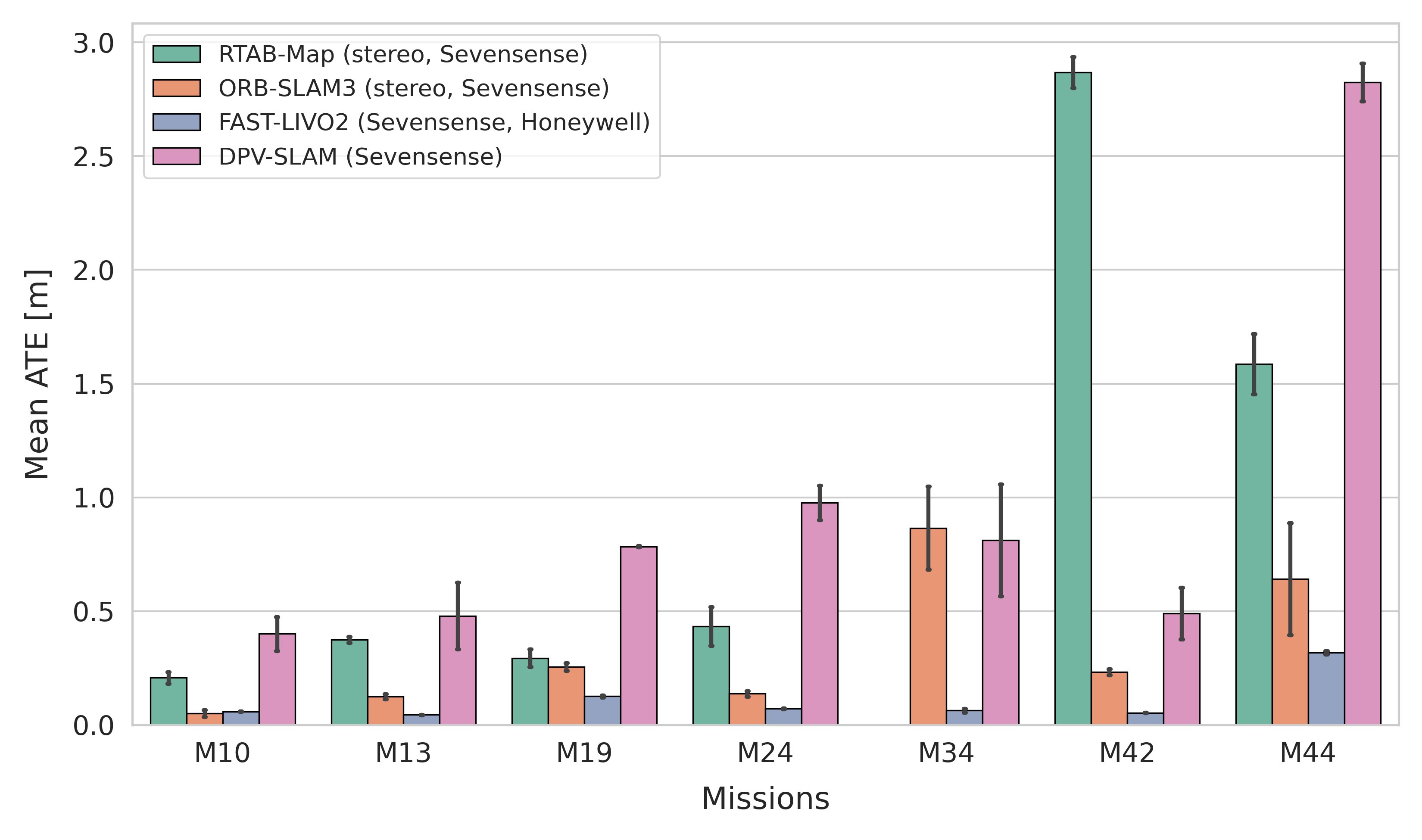}
    % \caption{Mean ATE of the best-performing sensor configurations across missions}
    % \caption{Mean Absolute Trajectory Error (ATE) for the top-performing sensor configurations. While the LiDAR-visual-inertial framework (FAST-LIVO2) achieves bounded drift across all environments, purely vision-centric methods exhibit localized failures triggered by low-texture subterranean spaces (M34), aggressive indoor looping (M42), and repetitive outdoor structures (M44).}
    \caption{Mean ATE for top-performing configurations. The LiDAR-visual-inertial pipeline (FAST-LIVO2) maintains bounded drift globally, whereas purely visual methods experience localized failures triggered by low texture (\textit{M34}), aggressive indoor turns (\textit{M42}), and structural repetition (\textit{M44}).}
    \label{fig:ate_best_configs}
\end{figure}

\begin{table}[htb]
\centering
\caption{Summary of mean accuracy metrics and resource usage across SLAM configurations reporting CPU load [\%] and RAM footprint [GB].}
\label{tab:best_config}
\small
\setlength{\tabcolsep}{4pt} % Standard padding for clean readability
\footnotesize
\begin{tabularx}{\linewidth}{l Y Y Y Y}
\toprule
\textbf{Method (sensor configuration)} & \textbf{ATE} & \textbf{RPE} & \textbf{CPU} & \textbf{RAM} \\
\midrule
ORB-SLAM3 (stereo, Sevensense)      & \nd 0.33 & \fst 0.02 & 289.98 & \nd 1.16 \\
RTAB-Map (stereo, Sevensense)          & \rd 0.96 & 0.10 & \nd 115.67 & \fst 0.75 \\
FAST-LIVO2 (Sevensense, Honeywell) & \fst 0.11 & \nd 0.04 & \rd 124.52 & \rd 3.91 \\
DPV-SLAM (Sevensense)               & 0.97 & \rd 0.06 & \fst 82.26 & 4.41 \\
\bottomrule
\end{tabularx}

\vspace{-0.75em}

\end{table}

The best configuration of each method across the evaluated sensor configurations is listed in Table \ref{tab:best_config}. These are the stereo mode using the global shutter camera for ORB-SLAM3 and RTAB-Map, the global shutter camera and high-end Honeywell IMU for FAST-LIVO2 and using the global shutter camera for DPV-SLAM.
Here, FAST-LIVO2 achieves the best performance in terms of ATE while ORB-SLAM3 minimizes local drift to capture the lowest RPE. 
From a resource perspective, DPV-SLAM exhibits the lowest CPU load while RTAB-Map requires the least memory footprint.

Further analysis of these optimal configurations across different environments is visualized in Fig. \ref{fig:ate_best_configs}. 
Notably, the RTAB-Map stereo configuration fails completely during mission \textit{M34} due to repeating turn maneuvers within a textureless subterranean landscape. 
Additionally, missions \textit{M42} and \textit{M44} prove exceptionally demanding for several visual frameworks. 
Mission \textit{M42} introduces severe indoor visual degradations including dynamic obstructions and quick reverse loop maneuvers. 
Meanwhile, mission \textit{M44} presents an outdoor route along a railway track featuring loose gravel and highly repetitive environmental structures. 
DPV-SLAM struggles to maintain tracking consistency during this railway sequence and displays a relatively high performance variance across a majority of the evaluation sequences, with the sole exception of the rough alpine terrain in mission \textit{M19}. 
Similarly, ORB-SLAM3 demonstrates a highly competitive localization profile but experiences elevated error distributions during the extreme illumination shifts of mission \textit{M34} and the structural repetitions of mission \textit{M44}. 
In contrast, FAST-LIVO2 leverages its active range sensor to anchor its filter backend, yielding exceptional tracking consistency and minimal variance across all evaluations, exhibiting only a minor increase in error during the rough terrain and dynamic movements of mission \textit{M44}.

\section{CONCLUSION}
\label{sec:conc}

In this work, we analyzed the influence of varying sensor configurations for state estimation on robustness, accuracy and resource utilization with quadruped robots using the GrandTour dataset. 
We showed that sensor configurations have substantial influence on robustness and accuracy.  
In general, stereo configurations outperformed monocular and RGB-D configurations. 
Furthermore, global shutter configurations consistently achieved better results in mean ATE and RPE across all tested sensor configurations than rolling shutter configurations. 
In addition, the high-end Honeywell IMU improved the ATE and RPE for the multi-modal, tightly integrated FAST-LIVO2 whereas ORB-SLAM3 exhibited reduced robustness for inertial configurations.
These findings can play an important role in the design of robot hardware and custom sensor payloads. Summarized, stereo configurations using global shutter cameras without inertial integration should be used for vision-centric methods to ensure robust localization accuracy during dynamic legged locomotion. For LiDAR-based methods, the addition of tactical-grade IMUs can improve accuracy compared to mid-range IMUs.
% Using the best sensor configurations, we compare mean ATE, RPE and resource utilization across all missions. 
% Here, FAST-LIVO2 achieves the best results for ATE and ORB-SLAM3 achieves the best results for RPE, while RTAB-Map achieves the lowest memory footprint and DPV-SLAM the lowest load.
%%%%%%%%%%%%%%%%%%%%%%%%%%%%%%%%%%%%%%%%%%%%%%%%%%%%%%%%%%%%%%%%%%%%%%%%%%%%%%%%

\bibliographystyle{IEEEtran}
\bibliography{bibliography}

\end{document}